\documentclass[sigconf]{acmart}

\AtBeginDocument{%
  }

\usepackage{graphicx}
\usepackage{multirow}

\setcopyright{acmcopyright}
\copyrightyear{2018}
\acmYear{2018}
\acmDOI{XXXXXXX.XXXXXXX}

\acmConference[MM '23]{the 31st ACM International Conference on Multimedia}{October 29--November 3, 2023}{Ottawa, Canada}
\acmPrice{15.00}
\acmISBN{978-1-4503-8651-7/21/10}

\begin{document}

%%
%% The "title" command has an optional parameter,
%% allowing the author to define a "short title" to be used in page headers.
\title{Sketch Input Method Editor: A Comprehensive Dataset and Methodology for Systematic Input Recognition}

\author{Guangming Zhu}
\affiliation{%
	\institution{Xidian University}
	\city{Xi'an}
	\country{China}}
\email{gmzhu@xidian.edu.cn}

\author{Siyuan Wang}
\affiliation{%
	\institution{Xidian University}
	\city{Xi'an}
	\country{China}}
\email{siyuanwang@stu.xidian.edu.cn}

\author{Qing Cheng}
\affiliation{%
	\institution{National University of Defense Technology}
	\city{Changsha}
	\country{China}}
\email{sgggps@163.com}

\author{Kelong Wu}
\affiliation{%
	\institution{Xidian University}
	\city{Xi'an}
	\country{China}}
\email{22031212500@stu.xidian.edu.cn}

\author{Hao Li}
\affiliation{%
	\institution{Xidian University}
	\city{Xi'an}
	\country{China}}
\email{xdulihao@stu.xidian.edu.cn}

\author{Liang Zhang}
\affiliation{%
	\institution{Xidian University}
	\city{Xi'an}
	\country{China}}
\email{liangzhang@xidian.edu.cn}

%%
%% The "author" command and its associated commands are used to define
%% the authors and their affiliations.
%% Of note is the shared affiliation of the first two authors, and the
%% "authornote" and "authornotemark" commands
%% used to denote shared contribution to the research.
%\author{Guangming Zhu}
%\authornote{Both authors contributed equally to this research.}
%\orcid{1234-5678-9012}
%\authornotemark[1]
%\affiliation{%
%  \institution{Xidian University}
%  \city{Xi'an}
%  \state{Shaanxi}
%  \country{China}
%}
%\email{gmzhu@xidian.edu.cn}

%\author{John Smith}
%\affiliation{%
%  \institution{The Th{\o}rv{\"a}ld Group}
%  \streetaddress{1 Th{\o}rv{\"a}ld Circle}
%  \city{Hekla}
%  \country{Iceland}}
%\email{jsmith@affiliation.org}

%\author{Julius P. Kumquat}
%\affiliation{%
%  \institution{The Kumquat Consortium}
%  \city{New York}
%  \country{USA}}
%\email{jpkumquat@consortium.net}

%%
%% By default, the full list of authors will be used in the page
%% headers. Often, this list is too long, and will overlap
%% other information printed in the page headers. This command allows
%% the author to define a more concise list
%% of authors' names for this purpose.
\renewcommand{\shortauthors}{Anonymous et al.}

%%
%% The abstract is a short summary of the work to be presented in the
%% article.
\begin{abstract}
  With the recent surge in the use of touchscreen devices, free-hand sketching has emerged as a promising modality for human-computer interaction. While previous research has focused on tasks such as recognition, retrieval, and generation of familiar everyday objects, this study aims to create a Sketch Input Method Editor (SketchIME) specifically designed for a professional C4I system. Within this system, sketches are utilized as low-fidelity prototypes for recommending standardized symbols in the creation of comprehensive situation maps. This paper also presents a systematic dataset comprising 374 specialized sketch types, and proposes a simultaneous recognition and segmentation architecture with multilevel supervision between recognition and segmentation to improve performance and enhance interpretability. By incorporating few-shot domain adaptation and class-incremental learning, the network's ability to adapt to new users and extend to new task-specific classes is significantly enhanced. Results from experiments conducted on both the proposed dataset and the SPG dataset illustrate the superior performance of the proposed architecture. Our dataset and code are publicly available at \url{https://github.com/Anony517/SketchIME}.
\end{abstract}

%%
%% The code below is generated by the tool at http://dl.acm.org/ccs.cfm.
%% Please copy and paste the code instead of the example below.
%%



%%
%% Keywords. The author(s) should pick words that accurately describe
%% the work being presented. Separate the keywords with commas.
%% A "teaser" image appears between the author and affiliation
%% information and the body of the document, and typically spans the
%% page.
%\begin{teaserfigure}
%	\includegraphics[width=\textwidth]{some sketches}
%	\caption{Examples of the sketched symbols in our dataset.}
%	\Description{some sketch categories}
%	\label{fig:some_sketch_categories}
%\end{teaserfigure}

%\received{20 February 2007}
%\received[revised]{12 March 2009}
%\received[accepted]{5 June 2009}

%%
%% This command processes the author and affiliation and title
%% information and builds the first part of the formatted document.
\maketitle

\section{Introduction}
Free-hand sketching is a communication modality that transcends barriers and connects human societies \cite{xu2022deep}. Despite its abstract and concise nature, it can be illustrative, making it useful in various scenarios such as communication and design. It has been extensively studied in fields such as computer vision \cite{wang2015sketch, yu2017sketch, ha2018neural}, computer graphics \cite{sangkloy2016sketchy}, human-computer interaction \cite{huang2019swire, gasques2019you} fields. Many related tasks have also been researched, including recognition \cite{yu2017sketch}, retrieval \cite{choi2019sketchhelper}, generation \cite{das2021cloud2curve}, grouping \cite{li2018universal}, segmentation \cite{wu2018sketchsegnet}, and sketch-based image retrieval \cite{hu2013performance}.

\begin{figure*}
	\includegraphics[width=\textwidth]{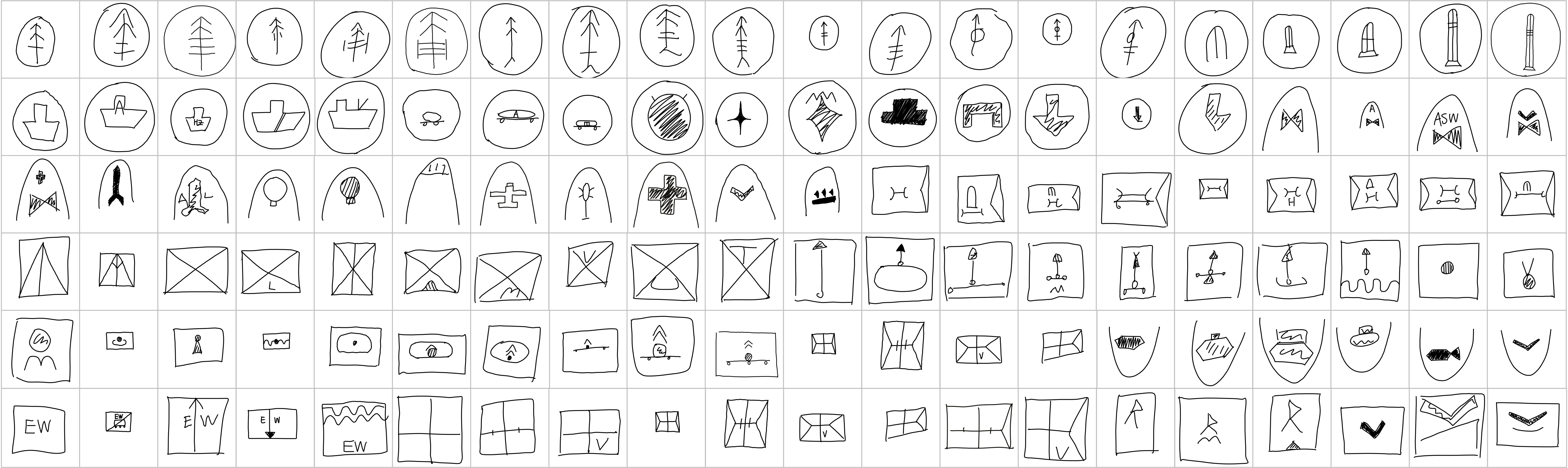}
	\caption{Examples of the sketched symbols in our dataset. These sketches have different sketching styles, large diversity on stroke length and count, and subtle differences between some categories.}
	\label{fig:some_sketch_categories}
\end{figure*}

In the era after WIMP (Windows, Icons, Menus, Pointer), there is a growing interest in using multi-touch, hand gesture, speech, and gaze analysis to achieve more intelligent and natural human-computer interaction \cite{bulling2012gaze}.Free-hand sketching is another natural form of interaction that has been used since ancient times and is even learned by children before they start writing. Its universal nature allows it to transcend language barriers \cite{xu2022deep}. While the nature of free-hand sketching limits its usage in standardized WIMP-based Graphical User Interfaces (GUIs), the increasing prevalence of touchscreen devices has made it a highly promising input modality that allows users to quickly sketch their concepts as low fidelity prototypes\cite{suleri2019eve}. 

Imagine drawing a flow chart using software like Visio or PowerPoint. You would need to select or drag the operational symbols from the toolbars to the drawing area before adjusting and editing them. However, this process can be time-consuming when the hierarchical toolbars contain a large number of operational symbols, especially for systems like the professional Command, Control, Communications, Computer, and Intelligence (C4I) system that contains over a thousand types of operational symbols. Our proposed solution is the Sketch Input Method Editor (SketchIME) system, which allows users to first sketch their desired symbols as low-fidelity prototypes. The system then automatically recognizes the symbols and their categories, replaces the sketches with corresponding standardized symbols at the correct positions, size, and orientation. This improves the overall efficiency of the interaction.

A SketchIME system should be designed in a way that enables it to recognize sketches and segment them into different semantic components. This will allow standardized symbols to be selected as replacements and adjusted according to the control points obtained from the segmentation results. Additionally, given the nature of free-hand sketch and the diversity of sketching styles among users, a well-designed SketchIME should be able to adapt to various styles online. Moreover, the system's ability to extend to new task-specific classes can make it even more attractive to users.

Using the aforementioned motivation and analysis, we collected and annotated a sketch dataset based on the public standard of the APP-6(B) joint military symbolism for C4I systems \footnote{https://www.scribd.com/document/314367702/APP-6-B-Joint-Symbology-pdf}. This standard outlines common operational symbols and their display/plotting details. We recruited a total of 67 participants to sketch the selected 374 types of symbols, as shown in Fig. \ref{fig:some_sketch_categories}. This is the first large-scale and systematic dataset to feature sketch categories from a professional C4I system, as opposed to familiar everyday objects. Our proposed network is a simultaneous recognition and segmentation architecture, with the recognition stream providing supervision to the segmentation stream based on prior knowledge. The incorporation of multilevel supervision enhances the interpretability of the network. Additionally, the utilization of domain adaptation techniques enhances the network's ability to adapt to new personal sketching styles at the application level. In addition, we have also explored a few-shot class-incremental learning mechanism to improve the network's ability to extend to new task-specific classes.

%Based on the above motivation and analysis, a sketch dataset based on the public standard of the APP-6(B) joint military symbology for C4I systems \footnote{https://www.scribd.com/document/314367702/APP-6-B-Joint-Symbology-pdf} was collected and annotated. This standard provides common operational symbols and details of their display/plotting. Total 67 participants are recruited to sketch the selected 374 types of symbols, as displayed in Fig. \ref{fig:some_sketch_categories}. This is the first large-scale and systematic dataset whose sketch categories come from a professional C4I system and not from familiar everyday objects.
%A simultaneous recognition and segmentation network is proposed, whose recognition stream provides supervision on the segmentation stream based on prior knowledge. Multilevel supervision makes the network more interpretable. Besides, the use of a domain adaptation method enables the network's adaptability to new personal sketching styles at an application-level. Furthermore, a few-shot class-incremental learning mechanism is explored to enhance the network's extendibility to new task-specific classes. 

Our contributions are summarized as follows:
\begin{itemize}
	\item The SketchIME dataset is the first of its kind, consisting of a large and systematic collection of sketches from a professional C4I system, rather than from familiar everyday objects. It is also annotated for easy reference.
	\item Our proposed approach is a simultaneous recognition and segmentation network that improves performance by incorporating multilevel supervision between recognition and segmentation tasks, while also enhancing interpretability.
	\item The incorporation of few-shot domain adaptation and class-incremental learning enhances the network's practicality by improving its adaptability to new users and extendibility to new task-specific classes
\end{itemize}

\section{Related Works}
\subsection{Sketch Dataset}
This paper focuses on the uni-modal free-hand sketch datasets. The widely used sketch datasets for recognition are TU-Berlin \cite{eitz2012humans} and QuickDraw \cite{ha2018neural}, which contain 250 and 345 object categories, respectively. Based on the QuickDraw dataset, some sketch datasets for grouping or segmentation are constructed, such as SPG \cite{li2018universal} and SketchSeg-150K \cite{qi2019sketchsegnet+}. SketchSeg-10K \cite{wang2019spfusionnet} is another segmentation dataset, which consists of 10 categories and 24 semantic labels. These datasets contains a lot of familiar everyday objects. 

In this study, we construct a large-scale and systematic dataset whose sketch categories come from a professional C4I system. In contrast to the large inter-category differences in the reviewed datasets, the difference between two categories in our dataset may be a single short line. Therefore, effective recognition and segmentation methods must learn the subtle differences.

\subsection{Sketch Recognition and Segmentation}
Sketch recognition aims to predict the class label of a given sketch. Sketch-a-Net \cite{yu2015sketch} is the first Convolutional Neural Network (CNN) based sketch recognition model that beats the human performance. Since a sketch is plotted as a continuous sequence of stroke points, Recurrent Neural Network (RNN) based models are also proposed in the literature. Stroke vectors or CNN features of strokes are fed to these models for sketch recognition \cite{sarvadevabhatla2016enabling, xu2018sketchmate, jia2017sequential, he2017sketch, prabhu2018distribution}. In contrast to the above models which exploit either the static nature of sketches with CNNs or the temporal sequential property with RNNs, sketches are represented as multiple sparsely connected graphs in \cite{xu2021multigraph}, and the global and local geometric stroke structures as well as temporal information are simultaneously captured in the multi-graph transformer.

\begin{table}
	\caption{Comparison between uni-modal sketch datasets.  Our SketchIME dataset contains the largest number of classes for recognition and the largest number of semantic components for segmentation. ``seg'' denotes the segmentation annotation, and ``SC-Count'' denotes the type count of semantic components in each dataset.}
	\label{tab:datacomparison}
	\resizebox{\linewidth}{!}{
	\begin{tabular}{cccccc}
		\toprule
		\textbf{Datasets} & \textbf{Amount} & \textbf{Category} & \textbf{Stroke} & \textbf{Annotations} & \textbf{SC-Count}\\
		\midrule
		TU-Berlin\cite{eitz2012humans} & 20K & 250 & $\checkmark$ & class & - \\
		QuickDraw\cite{ha2018neural} & 50M+ & 345 & $\checkmark$ & class & -\\
		QuickDraw-5-step\cite{sketchhelper} & 38M+ & 345 & - & class & -\\
		SketchFix-160\cite{sketchfix160} & 3904 & 160 & $\checkmark$ & class & -\\
		COAD\cite{TIRKAZ20123926} & 620 & 20 & $\checkmark$ & class & -\\		
		SPG\cite{li2018universal} & 20K & 25 & $\checkmark$ & class, grouping & 86\\
		SketchSeg-150K\cite{qi2019sketchsegnet+} & 150K & 20 & $\checkmark$ & class, seg & 57\\
		SketchSeg-10K\cite{wang2019spfusionnet} & 10K & 10 & $\checkmark$ & class, seg & 24\\
		\midrule
		\textbf{SketchIME(Our)} & 56K & \textbf{374} & $\checkmark$ & class, seg & \textbf{139}\\
		\bottomrule
	\end{tabular}}
\end{table}

Compared to sketch recognition, segmentation is a more fine-grained analysis task. CNN, RNN, and Graph Convolutional Network (GCN) based models have been proposed for stroke-level analysis. CNN-based network in \cite{li2018fast} takes the binary image of a sketched object as input and produces a corresponding segmentation map with per-pixel labels. The SketchSegNet \cite{wu2018sketchsegnet} and SketchSegNet+ \cite{qi2019sketchsegnet+} work in a RNN-based Variational Auto Encoder (VAE) pipeline. SPFusionNet \cite{wang2019spfusionnet} uses late fusion of CNN-RNN branches to represent sketches for segmentation. SketchGNN \cite{yang2021sketchgnn} uses a convolutional Graph Neural Network (GNN) for semantic segmentation and uses a mixed pooling block to fuse the intra-stroke and inter-stroke features.

\subsection{Adaptability and Extendibility}
Users may have their own personal sketching styles, similar to handwriting. A trained model may not work well for a new user who has a different sketching style. This means that the training data and the new users' data have different underlying distributions. Domain Adaptation (DA) is therefore needed to reduce the effect of domain shift \cite{pan2009survey} for SketchIME applications. There are different types of DA techniques \cite{preciado2021evaluation}. For instance, DeepCORAL \cite{sun2016deep} is a DA method based on correlation alignment \cite{sun2017correlation}. The Deep Domain Confusion (DDC) \cite{tzeng2014deep} uses an additional domain confusion loss to learn a domain invariant representation. The Conditional Domain Adversarial Network (CDAN) and its variant CDAN+E \cite{long2018conditional} use an adversarial DA technique, motivated by the conditional generative adversarial networks \cite{mirza2014conditional}, and make use of the discriminative information obtained from a deep classifier network. Based on CDAN, DA-FSL \cite{zhao2021domain} further poses an additional challenge of DA with few training samples. These methods can be explored to ensure the adaptation ability of SketchIME.

The ability to incrementally learn new sketch classes is crucial for SketchIME, particularly in cases where additional task-specific sketches were not included in the original training. Users are not obliged to capture large data for the evolution of trained models. Therefore, it is crucial to apply Few-Shot Class-Incremental Learning (FSCIL) \cite{tao2020few} technique to SketchIME. FSCIL is widely studied for image classification. Zhang \cite{zhang2021few} proposed a Continually Evolved Classifier (CEC) to update the classifier only in each incremental session to avoid knowledge forgetting. Zhu \cite{zhu2021self} proposed an incremental prototype learning scheme that consists of a random episode selection strategy and a Self-Promoted Prototype Refinement (SPPR) mechanism. CEC \cite{zhang2021few} and SPPR \cite{zhu2021self} only consider extending the model in a single phase, while LIMIT \cite{zhou2022few} considers extension in multiple phases. In contrast to the above fake-task based training methods, C-FSCIL \cite{hersche2022constrained} relies on the hyper-dimensional embedding space to incrementally create maximally separable classes. The ForwArd Compatible Training (FACT) \cite{zhou2022forward} assigns virtual prototypes to squeeze embeddings of known classes and reserve for new ones.

\section{Dataset Construction}
\subsection{Category Selection}
Total 374 categories of symbols are selected from the public standard of the APP-6(B) joint military symbology for our sketch collection. Fig.\ref{fig:some_sketch_categories} displays some sketched symbols, and all the 374 standardized symbols and their sketches are shown in the supplementary materials. Our dataset is different from the existing ones in several ways: (\textbf{a}) The sketch categories come from a professional C4I system, (\textbf{b}) Only subtle differences exist between some sketch categories, such as a single short line, (\textbf{c}) Blanks in some sketches need to be filled, and this results in a large diversity on stroke length and count, (\textbf{d}) The subtle differences between sketch categories and the large diversity of different personal sketching styles bring challenges to universal recognition and segmentation networks.  

\begin{figure}
	\centering
	\includegraphics[width=1\linewidth]{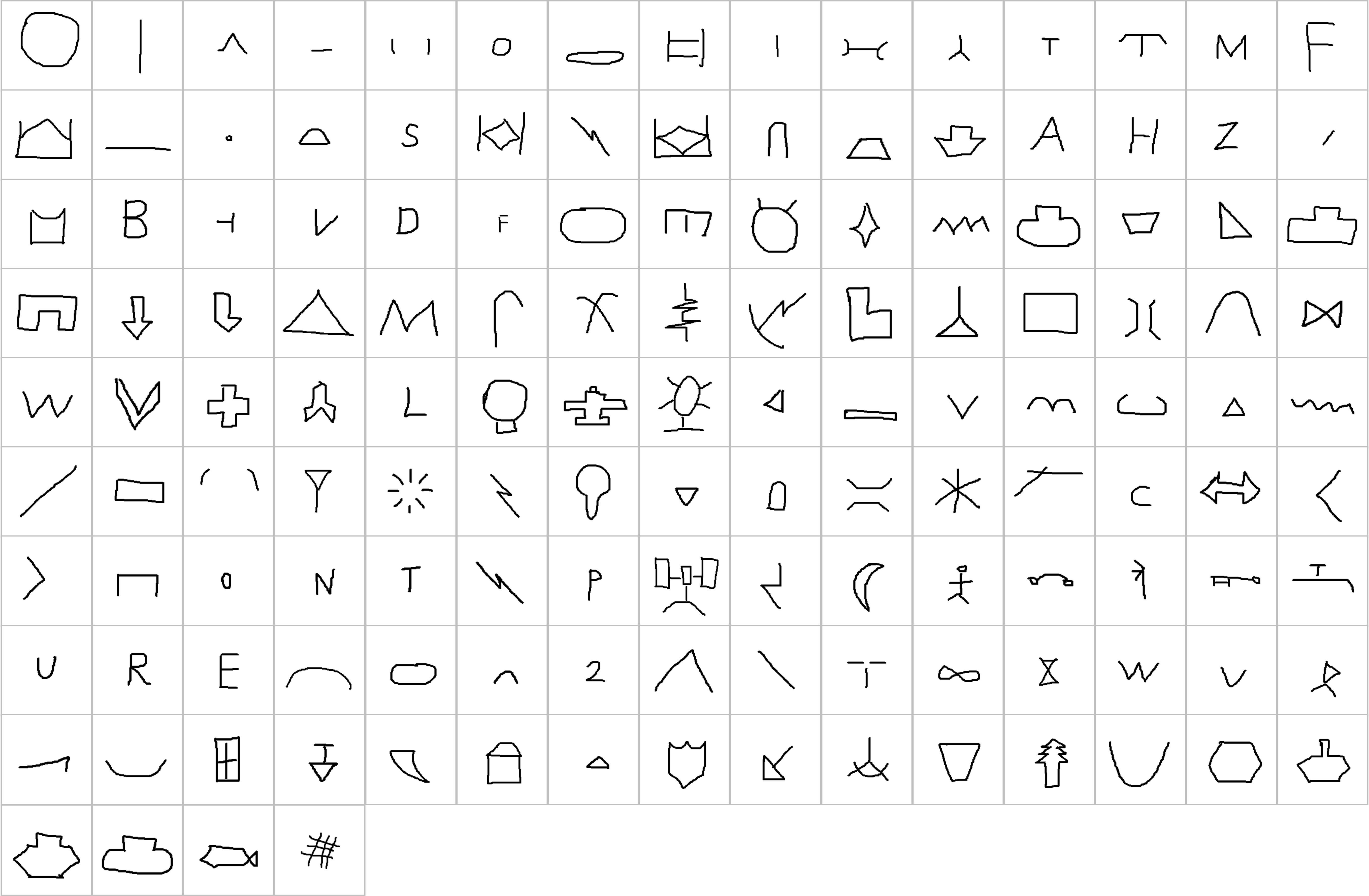}
	\caption{Semantic components of sketches in our dataset. The last semantic component denotes blank filling which may be done arbitrarily and result in a large diversity on stroke length and count. }
	\label{fig:all_components}
\end{figure}

\subsection{Data Collection and Annotation}
Total 67 participants were recruited to sketch the 374 categories of symbols using the Concepts Application \footnote{https://concepts.app/en/}. Phones or Pads with the Android or iOS operating systems were used to sketch. A touch pen was either used or not by participants, according to their personal preference. This also results in different sketching styles. The Scalable Vector Graphics (SVG) format was used to store sketches. 

Besides the class labels for each sketch sample, semantic labels are assigned to strokes of sketches of all the categories. The semantic labels are obtained according to the principle, ``\textit{If the increase or decrease of a stroke (or multiple consecutive strokes) will change one sketch's category, or a stroke (or multiple consecutive strokes) constitutes some class-specific component, the stroke (or strokes) is viewed as a semantic component.}'' It means that different categories may share semantic components. We numbered the semantic components of all categories uniformly and obtained 139 semantic labels, as displayed in Fig. \ref{fig:all_components}. Our work aims to recognize the categories and semantic components simultaneously. This is different from the grouping or segmentation tasks \cite{li2018universal,yang2021sketchgnn}. 

\subsection{Data Statistics}
Total 209K samples of the 374 categories were drawn. The publicly released version of the SketchIME dataset contains 56K samples, sketched by 8 participants for training and 10 participants for testing. Different participants may draw the same symbol using different stroke orders. Each participant only drew no more than 187 categories, and at most 25 samples were drawn for each category. Table \ref{tab:datacomparison} illustrates the comparison of our SketchIME dataset and popular uni-modal sketch recognition and segmentation datasets. For different specific tasks, the dataset is divided into different training and testing sets, and the numbers of sketches in the training and testing sets are illustrated in Table \ref{tab:datasets}. Our SketchIME dataset is constructed and annotated from scratch, and contains the largest number of classes for recognition (i.e., 374 categories) and segmentation (i.e., 374 categories and 139 semantic labels).

\section{Methods}
\label{sec:methods}
The black-box properties of deep learning restrict its applications in professional fields. The interpretability of neural networks can make them reliable. In this study, we propose a simultaneous recognition and segmentation network, whose multilevel supervision between recognition and segmentation based on prior knowledge makes the inference more interpretable. In addition, the use of few-shot domain adaptation improves the network's adaptability to new users' sketch styles, and the use of few-shot class-incremental learning enhance its extendibility to new task-specific classes. These features enhance the network's practicality.

\subsection{Recognition and Segmentation}
Our study suggests that, CNNs are effective for sketch-level recognition on images, and GNNs are better for stroke-level segmentation on SVG format. CNNs can learn sketch-level features effectively, while GNNs can take full use of stroke-level and point-level relationships between stroke points for segmentation. Furthermore, the prior knowledge about semantic components which should be contained in each sketch category, can also be used. Therefore, a simultaneous Sketch Recognition and Segmentation network (SketchRecSeg) is proposed in this study, whose recognition stream supervises the segmentation stream based on the prior knowledge. This allows the network to not only inference the sketch category, but also explain how the category is predicted based on the semantic component recognition.

\begin{figure}
	\centering
	\includegraphics[width=1\linewidth]{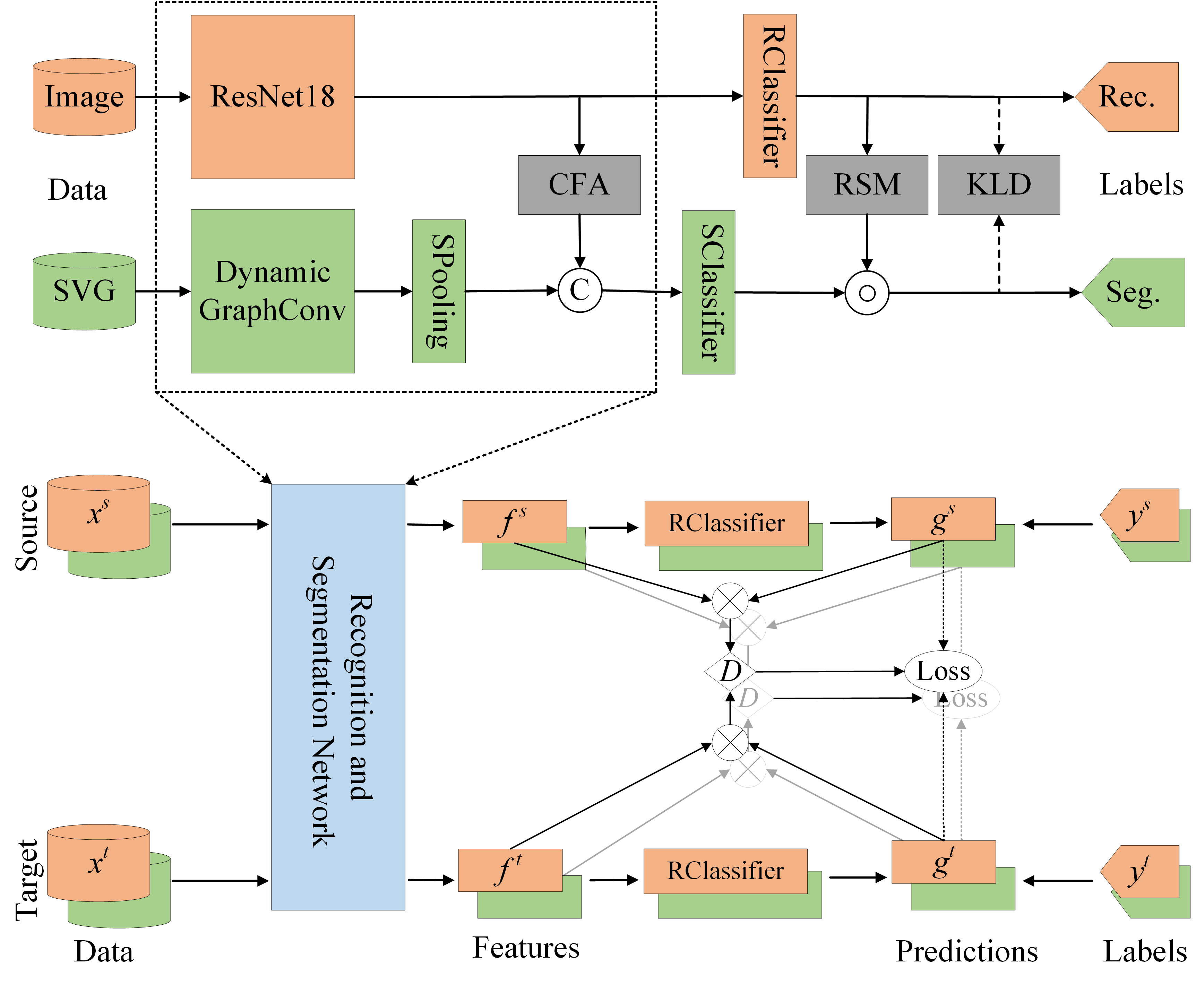}
	\caption{The proposed simultaneous sketch recognition and segmentation architecture. The top half denotes the two-stream architecture for simultaneous recognition and segmentation. The bottom half denotes the domain adaptation architecture using conditional domain adversarial. RClassifier and SClassifier denotes classifiers of categories and semantic components. The SPooling, CFA and RSM denotes the Stroke-level Pooling, the CNN Feature Augmentation, and the Recognition Supervision Module, respectively. KLD denotes the Kullback–Leibler divergence loss. $\copyright$ denotes feature concatenation operation, and $\circledcirc$ denotes Hadamard product.  $f^s$ and $f^t$ denote the domain-specific feature representation, $g^s$ and $g^t$ denote the classifier predictions, \textit{D} denotes the domain discriminator, and $\otimes$ denotes multilinear map operation.}
	\label{fig:sketchrecseg_network}
\end{figure}

The two-stream architecture of our SketchRecSeg is illustrated in Fig.\ref{fig:sketchrecseg_network}. The recognition stream uses a CNN network for feature learning and sketches are fed as images to this network. The segmentation stream uses a dynamic graph convolution unit and takes stroke points of SVG-format sketches as input. The features of the CNN network are taken as sketch-level features to augment the node features of the GNN block for segmentation. The recognition stream provides multilevel supervision (i.e., feature-level supervision using CFA, prediction-level supervision using RSM, and loss-level supervision using KLD) on segmentation. Besides, a domain adaptation mechanism \cite{long2018conditional} is equipped to support SketchRecSeg's adaptability to new users. 

\textbf{Feature Extraction.} We use ResNet18 \cite{he2016deep} to learn sketch-level features, and use the dynamic graph convolution unit \cite{yang2021sketchgnn} to learn point-level features. ResNet18 takes three-channel images transformed from the original SVG sketches as input, and outputs 128-dimensional sketch-level features by stacking an extra linear layer. The dynamic graph convolution units take a graph $\mathcal{G}=(\mathcal{V}, \mathcal{E})$ as input. $\mathcal{V}$ represents the $N$-point set $\mathcal{P}=\{p_i=(p^x_i,p^y_i)\}_{i=1,2,...,N}$, where $p^x_i$ and $p^y_i$ are the 2D absolute coordinates of point $i$ in strokes of a given sketch. $\mathcal{E}$ represents the edges that connect adjacent points in each single stroke. The graph convolution operation in \cite{wang2019dynamic} is used, and $\mathcal{E}$ is updated layer-by-layer using the Dilated k-NN \cite{li2019deepgcns}. The dynamic graph convolution unit contains four convolution blocks, each block outputs a 32-dimensional feature vector for each node, and a 128-dimensional feature vector is finally obtained by concatenating the four 32-dimensional vectors.

\textbf{Stroke-level Pooling (SPooling).} Each stroke contains several points, and all stroke points constitute the graph nodes. Although the dynamic graph convolution units have evolved the node features based on the updated neighborhood, using the node features alone still cannot achieve a satisfactory segmentation performance. The information about stroke and its relevant points can be obtained, and a stroke-level pooling strategy can be applied to augment the node features in stroke-level. Given the node features $\mathcal{F}_{g}=\{f_{g}^i\}_{i=1,2,...,N}$, SPooling can be implemented as:
\begin{equation}
f_{s}^j=\max\limits_{i\in \mathcal{V}_j}\mbox{MLP}(f_{g}^i),
\end{equation}
\begin{equation}
f_{sg}^i=\mbox{concat}_{i\in \mathcal{V}_j}(f_{s}^j, f_{g}^i),
\end{equation}
where $\mathcal{V}_j$ denotes the node subset belonging to stroke $j$, $f_{s}^j$ and $f_{sg}^i$ denote the stroke-level and the augmented features, respectively.

\textbf{CNN Feature Augmentation (CFA).} A simple global pooling on all graph nodes cannot provide effective sketch-level features. In contrast, the features of ResNet18 used for sketch recognition are the ideal representation of the whole sketch. Therefore, given the CNN feature $f_c$, the node features can be augmented further as:
\begin{equation}
f_{csg}^i=\mbox{concat}_{i\in \mathcal{V}}(f_{c}, f_{sg}^i).
\end{equation}

\textbf{Recognition Supervision Module (RSM).} This RSM module aims to use the sketch recognition probability to supervise the segmentation task. If a sketch is recognized belonging to some specific category, the semantic components making up the sketch category should also be predicted with high probability for segmentation task. Given the prior knowledge about semantic components, which sketch category contains, a translation matrix $\mathbf{C}_{r2s}\in\mathbb{R}^{C_R \times C_S}$ is constructed where $c_{ij}=\gamma_r$, when the $i$-th sketch category contains the $j$-th semantic component and otherwise $c_{ij}=1-\gamma_r$. $C_R$ and $C_S$ are the counts of sketch categories and semantic components. Given the sketch prediction probability $P_r \in \mathbb{R}^{1 \times C_R}$ and the indices $Idx_P$ of top-$k$ of $P_r$,  a probability vector that enhances the correct prediction of semantic components and suppresses the wrong prediction can be calculated as:
\begin{equation}
\Gamma_{sc}^r = MaxPooling(index\_select(\mathbf{C}_{r2s}, Idx_P)).
\end{equation}
Then, given the output $P_s \in \mathbb{R}^{N \times C_S}$ of the semantic component classifier in segmentation stream, the final semantic component prediction for point $i$ can be calculated as:
\begin{equation}
P_{s}^i = P_s^i\circ \Gamma_{sc}^r,
\end{equation}
where $\circ$ denotes the Hadamard product. Specifically, Eq. (5) is only applied when the confidence level of the category recognition exceeds a threshold. 

\textbf{Kullback-Leibler Divergence (KLD) Loss.} The total loss of SketchRecSeg can be represented as: 
\begin{equation}
L=L_s + \lambda_1 L_r + \lambda_2 L_{kl},
\end{equation}
where $L_s$ is the point-level segmentation loss, $L_r$ is the recognition loss, $\lambda_1$ is the coefficient to balance recognition and segmentation, and $\lambda_2$ is an indicator of whether $L_{kl}$ is used or not. $L_s$ and $L_r$ use cross-entropy loss. $L_{kl}$ is the KL-divergence loss that represents the recognition's supervision on segmentation, and can be denoted as:
\begin{equation}
L_{kl} = \mbox{KL}[P_{sc}^r||P_{sc}^s],
\end{equation}
where $P_{sc}^r$ and $P_{sc}^s$ denote the probability distributions of semantic components derived from the recognition and segmentation results, respectively. $P_{sc}^r$ can be calculated as:
\begin{equation}
P_{sc}^r = P_r*\mathbf{C}_{r2s}.
\end{equation}
The stroke-level semantic component prediction can be obtained by pooling on the point-level prediction results, and it also represents how much each semantic component should exist in this sketch. Therefore, $P_{sc}^s$ can be calculated as: 
\begin{equation}
P_{sc}^s = \max\limits_{j=1,...,J}(\frac{1}{|\mathcal{V}_j|}\sum\limits_{i\in \mathcal{V}_j}P_s^i),
\end{equation}
where $J$ is the stroke count of one sketch.

\textbf{Domain Adaptation (DA).} Users may have their own personal and preferred sketching styles. A trained model may not work well for a new user who has a different sketching style. This is because of the domain shift between the training data (source domain) and the new user's data (target domain). If using few samples of new users to fine-tune the trained model can improve the performance significantly, it can enhance SketchIME's practicality. It is reasonable to assume that few samples (e.g., empirically selected five per category) in the large-scale training data can be used as the source domain data for fine-tuning our model. These samples can be considered as typical examples of each category for user reference. Similar to using common input method editors, when users are using SketchIME, they will select the right symbol from the categories recommended by the recognition results. This process provides some labeled data as the target domain data. Given a small training data and new user's data, few-shot domain adaptation can enhance SketchIME's practicality. 

CDAN \cite{long2018conditional} can be used to enhance the network's adaptability to new users. Given the source classifier $G$ (including the feature extractor and the classifier in each stream of SketchRecSeg) and the domain discriminator $D$ (for each stream), the optimization problem of supervised domain adaptation for the recognition stream (or the segmentation stream) can be denoted as:
\begin{equation}
\min\limits_{G}E(G)-\lambda E(D,G) ~and~ \min\limits_{D}E(D,G),
\label{eq:cdan1}
\end{equation}
\begin{equation}
\begin{split}
E(G)=\mathbb{E}_{(x_i^s,y_i^s)\sim D_s}L(G(x_i^s), y_i^s) \\
+\mathbb{E}_{(x_j^t,y_j^t)\sim D_t}L(G(x_j^t), y_j^t),
\end{split}
\end{equation}
\begin{equation}
\begin{split}
E(D,G)=-\mathbb{E}_{x_i^s\sim D_s}log[D(f_i^s,g_i^s)]\\
-\mathbb{E}_{x_j^t\sim D_t}log[1-D(f_j^t,g_j^t)].
\end{split}
\label{eq:cdan3}
\end{equation}

\noindent where $f_i^s$ and $f_j^t$ are the feature representation of the CNN (or GNN) stream, $g_i^s$ and $g_j^t$ are the classifier predictions, $x_i^s$ and $x_j^t$ are sketch images (or strokes), $y_i^s$ and $y_j^t$ are ground-truth category (or semantic component) labels, and $D_s$ and $D_t$ are the source (few training data) and target (new users' data) domains.

\subsection{Extendibility Enhancement}
In a professional C4I system, there are more than one thousand types of sketches. Each sketch consists of basic semantic components, and new task-specific sketch categories (even new semantic components) may be derived from combinations of multiple semantic components. It means that, in order to enhance the SketchRecSeg network's forward compatibility to ensure its extendibility, we can pre-assign several virtual prototypes in the embedding space, and treat them as `virtual classes'. Therefore, the ForwArd Compatible Training (FACT) \cite{zhou2022forward}, which assigns virtual prototypes to squeeze the embedding of known classes and reserves for new ones, can be used for class-incremental learning.

Figure \ref{fig:sketchrecseg_network} shows that we need forward compatible training for recognition and segmentation streams simultaneously. The loss for each stream can be given as $L=L_v+L_f$:

\begin{equation}
L_v(\mathbf{x},y)={l}(f_v(\mathbf{x}),y)+\gamma {l}(\text{Mask}(f_v(\mathbf{x}),y),\hat{y}),
\end{equation}
\begin{equation}
L_f(\mathbf{z},y)={l}(f_v(\mathbf{z}),\hat{y})+\gamma {l}(\text{Mask}(f_v(\mathbf{z}),\hat{y}),\hat{\hat{y}}),
\end{equation}
\begin{equation}
\text{Mask}(f_v(\mathbf{x}),y)=f_v(\mathbf{x})\otimes(\mathbf{1}-\text{OneHot}(y)),
\end{equation}

\noindent where $\mathbf{x}$ and $y$ are the data and label of known classes, $l$ is the cross-entropy loss, $\gamma$ is the trade-off parameter, $\mathbf{1}$ is an all-ones vector, and $\otimes$ is Hadamard product. The cosine classifier can be abbreviated as $f_v(\mathbf{x})=[W,P_v]^\top\phi(\mathbf{x})$, where $W$ and $P_v$ are the prototypes of known and virtual classes, and $\phi(\cdot)$ is the feature extractor. $P_v=[p_1,\cdots,p_V]\in R^{d\times V}$ denotes the prototypes of $V$ virtual classes. $\hat{y}=\text{argmax}_v p_v^\top\phi(\mathbf{x})+|Y_0|$ is the virtual class with maximum logit, acting as the pseudo label, where $|Y_0|$ is the class  count of known classes. $\mathbf{z}$ is virtual instance obtained by fusing the embeddings of two instances by manifold mixup. $\hat{\hat{y}}=\text{argmax}_kw_k^\top\mathbf{z}$ is the pseudo label among current known classes. The hyper-parameters have same values as in \cite{zhou2022forward}. The forward compatible training is implemented simultaneously for incremental learning of new classes and semantic components.

\section{Experimental Studies}
\label{sec:experiment}
\subsection{Datasets}
Three datasets are constructed from the collected data, as illustrated in Table \ref{tab:datasets}. \textbf{\textit{SketchIME-SRS}} contains data of 8 participants for training and data of another 10 participants for testing. On the one hand, SketchIME-SRS is used for common evaluations based on the train and test subset. On the other hand, one or two samples from the 10 participants in the test subset can also be used for domain adaptation training when SketchRecSeg is equipped with DA, and another five samples from each category sketched by 10 participants are used for testing.

\textbf{\textit{SketchIME-CIL1}} and \textbf{\textit{SketchIME-CIL2}} are constructed for the extendibility evaluation. The former is used for the situation where both new sketch categories and new semantic components are expected to emerge in incremental sessions. The latter is used for the situation where all semantic components already exist in the base session and only new sketch categories emerge in incremental sessions. Each sketch category only has five samples in incremental sessions. We cannot design all the incremental sessions with the same count of new sketch categories and new semantic components. Therefore, 16 new sketch categories emerge in each incremental session first, and then the incremental semantic components in this session are extracted from the 16 new sketch categories. This results in different increment counts of semantic components in each incremental session. The details about the sketch categories and semantic components in base and incremental sessions are visualized in the supplementary materials.

\begin{table}
	\caption{Three datasets constructed from the collected data for
		three tasks. The numbers in square brackets denote the sketch category count, the semantic component category count, and the sample count in each subset, respectively. The numbers in the `Incremental' column denote the total counts of new sketches and semantic components emerging in incremental sessions.}
	\label{tab:datasets}
	\resizebox{\linewidth}{!}{
	\begin{tabular}{llll}
		\toprule
		\textbf{Datasets} & \textbf{Train/Base} & \textbf{Incremental} &  \textbf{Test} \\
		\midrule
		SketchIME-SRS & [374,139,36693] & - & [374,139,19781] \\
		SketchIME-CIL1 & [214,98,20904] & [160,41,800] & [374,139,19781] \\
		SketchIME-CIL2 & [214,139,21077] & [160,0,800] & [374,139,18986]\\
		\bottomrule
	\end{tabular}}
\end{table}

Besides, the SPG dataset \cite{li2018universal} is also used to verify the advantages of the proposed networks. SPG has 25 categories and 800 sketches per category. Similar to SketchGNN \cite{yang2021sketchgnn}, only 20 categories are used for evaluation. However, this study not only focuses on grouping strokes of one sketch into semantic parts, but also to recognize the semantic parts without knowing ahead the sketch's category label.

\subsection{Recognition and Segmentation}
\textbf{Training Details.} Image- and SVG-format sketches in SketchIME-SRS are fed to our SketchRecSeg network simultaneously. $N=300$ points are sampled from each sketch. The learning rate is initialized to $2\times 10^{-3}$ with a batch size of 256. Adam optimizer is used for optimization. Total 100 epochs are implemented. Since the segmentation stream has $N=300$ prediction outputs, $\lambda_1$ is set to 150 to balance recognition and segmentation losses. $\lambda_2$ is set to 1 or 0 depending on whether KL-divergence loss is used or not. The two-streams of SketchRecSeg are simultaneously trained from scratch. Our network is implemented in Pytorch and trained on a single NVIDIA GeForce GTX 3090.

\begin{table}
	\caption{Comparison results between the proposed SketchRecSeg and the state-of-the-art methods of sketch recognition and segmentation on SketchIME-SRS dataset.}
	\label{tab:sketchrecseg_result}
	\begin{tabular}{lccc}
		\toprule
		\textbf{Networks} & \textbf{P-Metric} & \textbf{C-Metric} & \textbf{Acc@1} \\
		\midrule
		Vision Transformer\cite{dosovitskiy2021image} & - & - & 31.66\\
		MultiGraph Transformer \cite{xu2021multigraph} & - & - & 32.63\\
		ResNet18 \cite{xu2022deep} & - & - & 59.10 \\
		\cline{1-4}	
		SketchSegNet \cite{wu2018sketchsegnet} & 44.86 & 25.44 & - \\		
		MultiGraph Transformer \cite{xu2021multigraph} & 51.09 & 27.81 & - \\
		SketchGNN \cite{yang2021sketchgnn} & 77.96 & 76.65 & - \\
		\cline{1-4}
		SketchRecSeg(CFA) & 84.18 & 83.56 & 76.48 \\
		SketchRecSeg(CFA+RSM) & 84.96 & 84.37 & 76.48 \\
		SketchRecSeg(CFA+KLD) & 85.45 & 84.87 & \textbf{77.21} \\
		SketchRecSeg(CFA+KLD+RSM) & \textbf{86.14} & \textbf{85.58} & \textbf{77.21} \\
		\cline{1-4}
		SketchRecSeg+DA1 & 94.82 & 95.01 & 89.50 \\
		SketchRecSeg+DA2 & 96.16 & 96.31 & 92.85 \\
		SketchRecSeg+DA5 & \textbf{97.01} & \textbf{97.13} & \textbf{96.21} \\
		\bottomrule
	\end{tabular}
\end{table}

\begin{figure}
	\centering
	\includegraphics[width=0.8\linewidth]{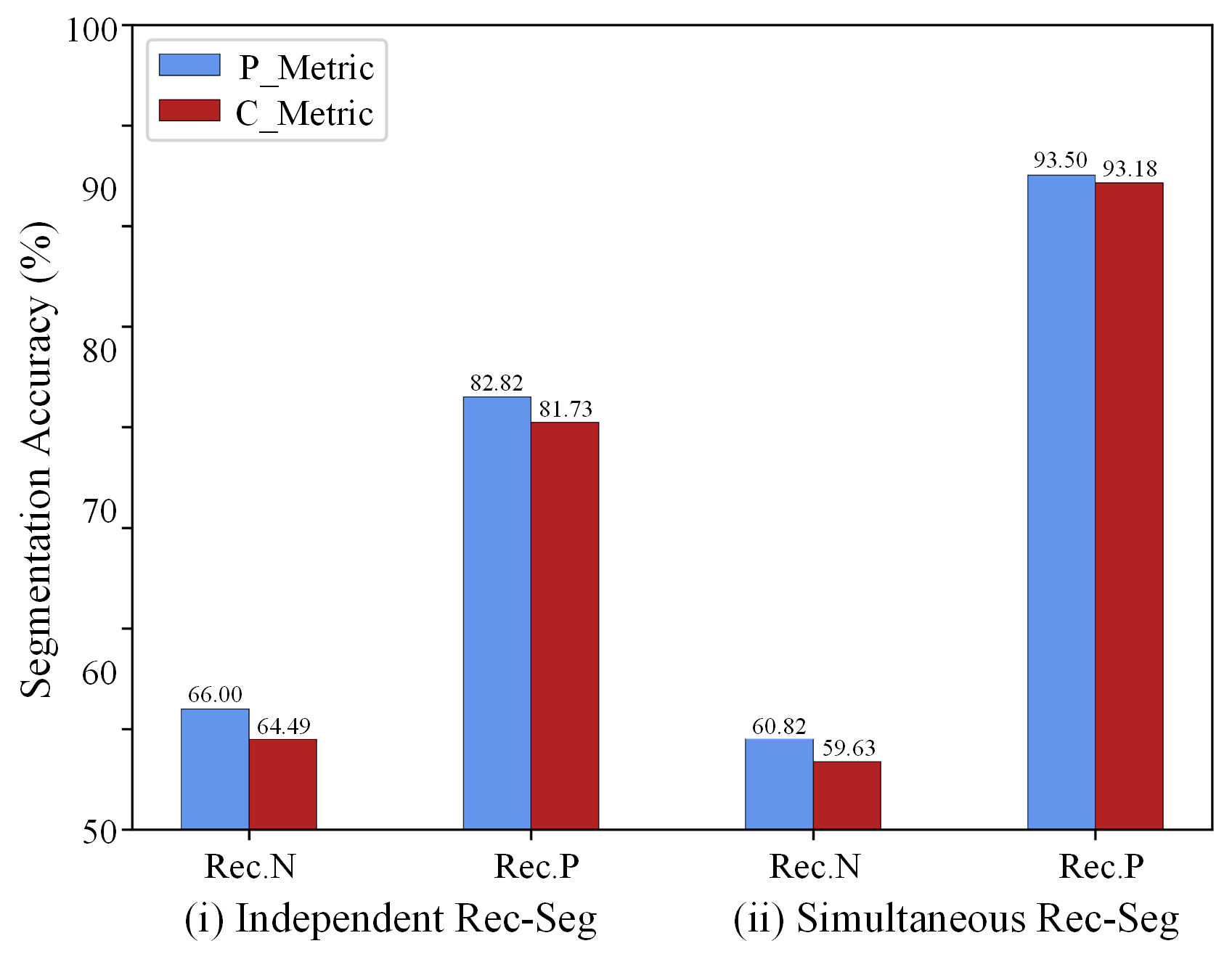}
	\caption{The interpretability analysis of the proposed SketchRecSeg. Rec.N and Rec.P denote the situations where the recognition network wrongly and correctly recognize sketches' categories, respectively. ``Independent Rec-Seg'' denotes the separate training of ResNet18 and SketchGNN networks. ``Simultaneous Rec-Seg'' denotes the proposed SketchRecSeg network. SketchRecSeg also achieves high segmentation performances when it correctly recognize sketches' categories. }
	\label{fig:mutual_effect}
\end{figure}

\textbf{Evaluation Metrics.} Sketch recognition is a typical classification problem, therefore Top-1 accuracy (\textbf{Acc@1}) is used as the evaluation metric. Similar to \cite{wu2018sketchsegnet} and \cite{yang2021sketchgnn}, two metrics are selected for segmentation evaluation, including (a) Point-based accuracy (\textbf{P\_Metric}) which evaluates the percentage of the correctly predicted stroke points in all sketches, and (b) Component-based accuracy (\textbf{C\_Metric}) which evaluates the percentage of the correctly predicted strokes. A stroke is considered to be correctly predicted if more than 75\% of its points are correctly predicted.

\textbf{Benchmark Methods.} To the best of our knowledge, the proposed SketchRecSeg is the first network that simultaneously recognizes and segments sketches. Therefore, some popular sketch recognition methods, e.g., CNN- \cite{xu2022deep} and Transformer-based \cite{xu2021multigraph} methods, are used for comparison. RNN-based recognition methods are not used in our experiments due to their sensitiveness to stroke orders. For segmentation, SketchSegNet \cite{wu2018sketchsegnet} and SketchGNN \cite{yang2021sketchgnn} are evaluated for comparison. Besides, the multigraph Transformer \cite{xu2021multigraph} can easily be modified to support sketch segmentation, therefore it is also evaluated.

\textbf{Performance Analysis.} Table \ref{tab:sketchrecseg_result} illustrates the evaluation results. \textit{\textbf{Firstly}}, the proposed simultaneous recognition and segmentation network achieves significant performance improvement, compared with the state-of-the-art methods. Vision Transformer \cite{dosovitskiy2021image} splits the image into multiple patches and uses linear layers to extract features from patches before inputting them to Transformer. The linear layers are not effective to learn two-dimensional features from images. Multigraph Transformer's positional embedding uses stroke orders \cite{xu2021multigraph}, therefore it cannot achieve good performance when samples in the test subset have different stroke orders compared to the ones in the train subset. SketchSegNet \cite{wu2018sketchsegnet} uses LSTM layers, and thus does not achieve good segmentation performance.

\textit{\textbf{Secondly}}, the proposed SketchRecSeg uses ResNet18 and dynamic graph convolutions for feature learning. It achieves significant improvement both for recognition and segmentation tasks, compared with ResNet18 for recognition and SketchGNN for segmentation separately, even when SketchRecSeg only uses feature-level supervision by CFA. This demonstrates that the two-stream network and the supervision can improve the recognition and segmentation performance simultaneously.

\textit{\textbf{Thirdly}}, the use of the prior knowledge by the RSM module and KLD loss brings considerable performance improvement on the segmentation stream, without affecting the recognition performance. The computational and memory consumption by RSM and KLD is very little. The multilevel supervision takes full use of the prior knowledge and the advantages of CNN and GNN, and thus results in better performance.

\textit{\textbf{Fourthly}}, Fig. \ref{fig:mutual_effect} illustrates our analysis about the mutual effects of recognition and segmentation. The left two groups show the segmentation performance of SketchGNN on the samples for which the recognition network (i.e., ResNet18) makes the wrong (i.e., shown in the Rec.N bars under the scenario of independent Rec-Seg) and the right (i.e., shown in the Rec.P bars under the scenario of independent Rec-Seg) predictions. Note that ResNet18 and SketchGNN are trained independently. The right two groups give the performance of our simultaneous recognition and segmentation network. It can be seen that our network has very high segmentation accuracy when it correctly recognizes the sketch categories. While, it performs poorly for segmentation task when it cannot recognize the sketch correctly. This implies that it can recognize the sketch correctly because it can correctly segment the sketch's semantic components, and vice versa. This makes the network reliable due to its interpretability. In conclusion, the proposed SketchRecSeg achieves better and balanced recognition and segmentation performance simultaneously. Simultaneous recognition and segmentation shows its superiority compared to independent implementations.

\textit{\textbf{Lastly}}, the bottom three rows of Table \ref{tab:sketchrecseg_result} show that supervised CDAN domain adaptation training using only one, two, five samples (corresponding to the cases of SketchRecSeg+DA1/DA2/DA5) per category of new users can significantly improve the performance to an application-level. It is reasonable to use few samples of the base training data for domain adaptation, since they can be viewed as the typical examples of each category for user reference. It is also reasonable to use few labeled samples of new users, since the labels can be obtained online when users select the right symbols from the recommended ones while using SketchIME. This provides us a few-shot domain adaptation method for personal sketching style learning, and ensures the adaptability of SketchIME. This also provides us a novel personal sketch style learning strategy to transfer the offline trained model to each new user' data for high-precision sketch recognition and segmentation.

\begin{figure}
	\centering
	\includegraphics[width=1\linewidth]{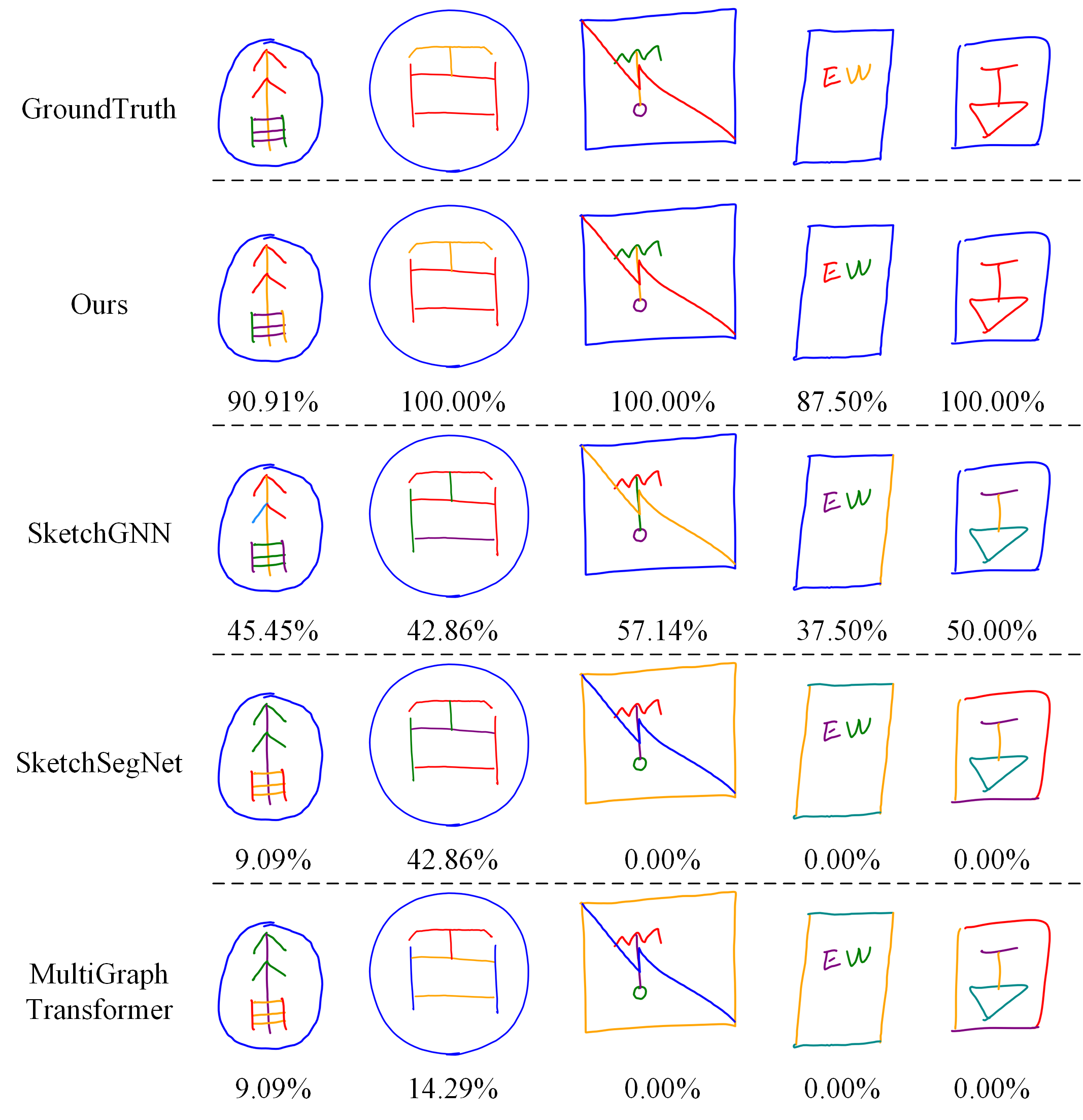}
	\caption{Visualization of some segmentation results on the SketchIME-SRS dataset. The numbers under the sketch samples denote the C-Metric values of each sample obtained by each method. It means correct segmentation only if the color of a stroke in each sample in the bottom-four rows matches the color of the same stroke in the top row. }
	\label{fig:seg_visualization}
\end{figure}

\begin{table}
	\caption{Comparison with state-of-the-art methods on SPG.}
	\label{tab:sketchrecseg_spg}
	\begin{tabular}{lccc}
		\toprule
		\textbf{Networks} & \textbf{P-Metric} & \textbf{C-Metric} & \textbf{Acc@1} \\
		\midrule
		Vision Transformer\cite{dosovitskiy2021image} & - & - & 76.21 \\
		BiGRU\cite{chung2014empirical} & - & - & 79.10 \\		
		ResNet18 \cite{xu2022deep} & - & - & 80.66 \\
		MultiGraph Transformer \cite{xu2021multigraph} & - & - & 91.05\\
		\cline{1-4}	
		SketchSegNet \cite{wu2018sketchsegnet} & 56.22 & 45.46 & - \\
		SketchGNN \cite{yang2021sketchgnn} & 91.26 & 87.86 & - \\
		\cline{1-4}
		SketchRecSeg(CFA) & 93.18 & 90.86 & 97.37 \\
		SketchRecSeg(CFA+KLD) & 93.83 & 91.25 & 97.47 \\
		SketchRecSeg(CFA+KLD+RSM) & 94.09 & 91.65 & 97.47 \\
		\bottomrule
	\end{tabular}
\end{table}

\begin{figure}
	\centering
	\includegraphics[width=1\linewidth]{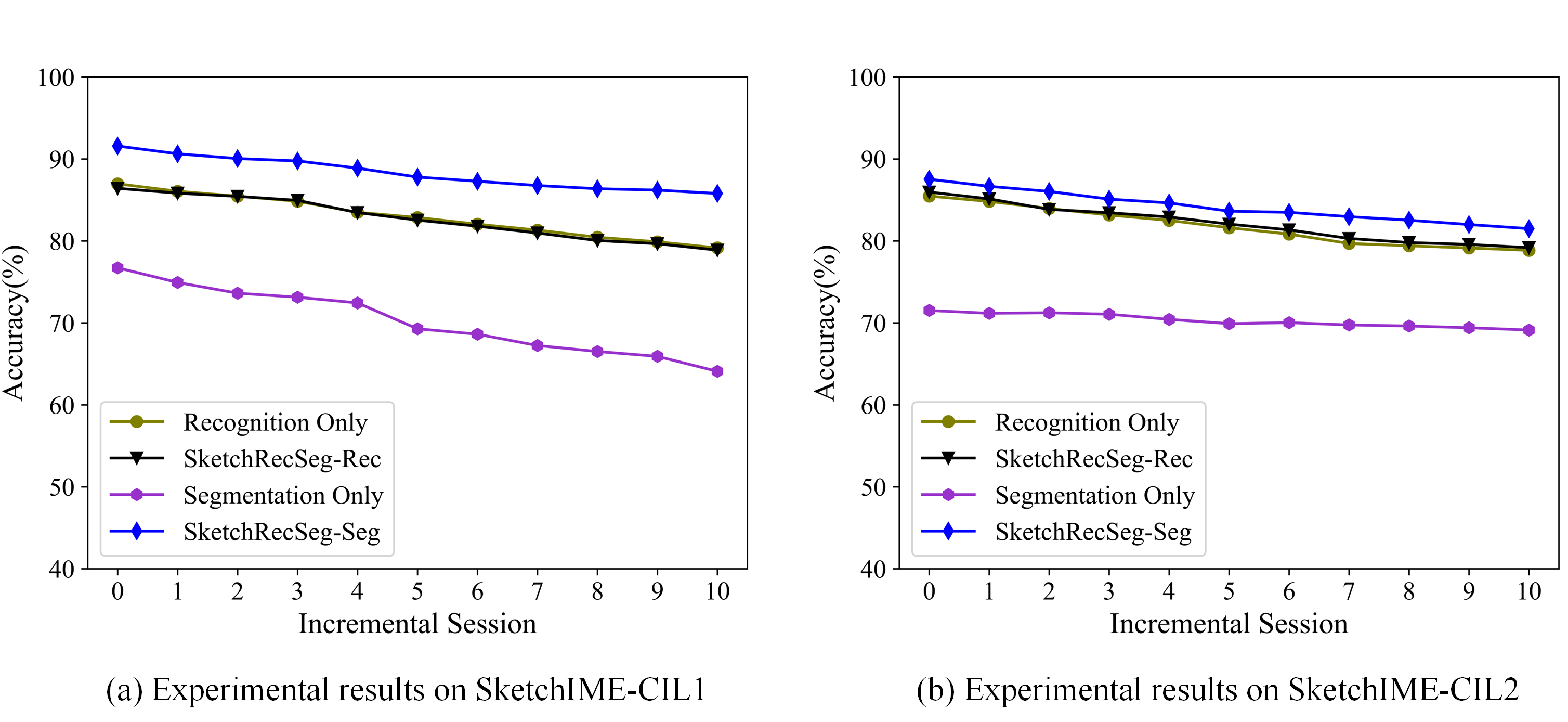}
	\caption{Experimental results of class-incremental learning. The proposed SketchRecSeg using forward compatible training has a good extendibility to new task-specific classes. }
	\label{fig:sketchcil_results}
\end{figure}

Figure \ref{fig:seg_visualization} shows some segmentation results on our SketchIME-SRS dataset. It visually shows the composition of a sketch's semantic component. It also demonstrates the advantages of the proposed SketchRecSeg.

\subsection{Extendibility Analysis}
\textbf{Experimental Setup.} Two experiments are performed on the SketchIME-CIL1 and SketchIME-CIL2 datasets. The former deals with the situation where both new sketch categories and new semantic components emerge in the incremental sessions, while the latter deals with the situation where only new sketch categories emerge in the incremental sessions. The forward compatible training \cite{zhou2022forward} of the recognition (i.e., ResNet18) and segmentation (i.e., SketchGNN) networks is performed separately as the baseline, and the results are illustrated in Fig. \ref{fig:sketchcil_results} as ``Recognition Only'' and ``Segmentation Only''. We also applied the forward compatible training \cite{zhou2022forward} to support our two-stream network to incrementally learn new sketch categories and new semantic components simultaneously, and the results on recognition and segmentation are illustrated in Fig. \ref{fig:sketchcil_results} as ``SketchRecSeg-Rec'' and ``SketchRecSeg-Seg'' respectively.

\textbf{Performance Analysis.} It can be seen from Fig.\ref{fig:sketchcil_results} that: \textbf{(a)} When dealing with the class-increment on both the sketch categories and semantic components, the joint optimization on the embedding spaces of categories and components can improve both recognition and segmentation performance (see Fig. \ref{fig:sketchcil_results}(a)), and the recognition's supervision on segmentation significantly improves the segmentation performance, \textbf{(b)} When only dealing with the class-increment on the sketch categories, the joint optimization on the embedding spaces of categories and components can force the segmentation stream to learn how the existing semantic components constitute new categories and benefit from the learning process, \textbf{(c)} Fig. \ref{fig:sketchcil_results} reports slight performance reduction for both recognition and segmentation in incremental sessions. This means that applying the forward compatible training strategy of FACT \cite{zhou2022forward} on our SketchRecSeg network can effectively enhance its extendibility to new task-specific sketches.  

~

In conclusion, given the analysis in Section 5.2 and 5.3, conclusions can be obtained:  \textbf{(a)} the two-stream architecture and the supervision between two streams can benefit recognition and segmentation simultaneously. \textbf{(b)} Applying the training strategies of few-shot domain adaptation and class-incremental learning can enhance the network's adaptability to new users and extendibility to new task-specific sketches, and enhance its practicality.  

\subsection{Comparison on Other Dataset}
Table \ref{tab:sketchrecseg_spg} illustrates the comparison results of our methods and state-of-the-art methods on the SPG dataset. SketchRecSeg outperforms both the recognition networks and the segmentation networks. Only subtle performance difference exists when SketchRecSeg uses a different supervision. This is because SketchRecSeg has already achieved excellent performance with CFA only.

\section{Discussion and Conclusion}
The traditional ``Search and Select'' input styles may be time consuming when frequent inputs are needed, especially when a large number of operational symbols are contained in hierarchical toolbars. Sketch input method editor (SketchIME) is a professional and promising system which can provide user-friendly and efficient input experience. Beyond the base recognition and segmentation functions, the adaptability and extendibility are also the key features to ensure SketchIME can be used by different users for various tasks. This study constructed datasets from scratch, proposed the simultaneous recognition and segmentation network, and explored the few-shot domain adaptation and class-incremental learning mechanism to improve the network's adaptability to new users and extendibility to new task-specific sketches. This study verifies the feasibility to use few samples to adapt to new users and tasks, and this paper provides systematic data and methods for SketchIME.  

For an excellent sketch input method editor, online recognition and recommendation are more important. The unused parts of our collected data contains different stroke orders of each sketch. These can be used for online sketch recognition and personal identification based on sketching styles and will be released in our future works.

%%
%% The acknowledgments section is defined using the "acks" environment
%% (and NOT an unnumbered section). This ensures the proper
%% identification of the section in the article metadata, and the
%% consistent spelling of the heading.

%\begin{acks}
%	thanks.
%\end{acks}

\clearpage

%%
%% The next two lines define the bibliography style to be used, and
%% the bibliography file.
\bibliographystyle{ACM-Reference-Format}
\bibliography{sample-base}

%%
%% If your work has an appendix, this is the place to put it.
\appendix

\end{document}